\documentclass[a4paper,twoside]{article}

\usepackage{epsfig}
\usepackage{subcaption}
\usepackage{calc}
\usepackage{amssymb}
\usepackage{amstext}
\usepackage{amsmath}
\usepackage{amsthm}
\usepackage{multicol}
\usepackage{pslatex}
\usepackage{apalike}
\usepackage{SCITEPRESS}     % Please add other packages that you may need BEFORE the SCITEPRESS.sty package.

\usepackage{times}
\usepackage{epsfig}
\usepackage{graphicx}
\usepackage{amssymb} 
\usepackage{amsmath}
\usepackage[linesnumbered,ruled]{algorithm2e}
\usepackage{array}
\usepackage{rotating}
\usepackage{multirow}
\usepackage{csquotes}
\usepackage{color}
\usepackage{booktabs}
\usepackage{cite}
\usepackage{amsmath,amssymb,amsfonts}
\usepackage{algorithmic}
\usepackage{graphicx}
\usepackage{url}
\usepackage{textcomp}
\usepackage[dvipsnames]{xcolor}
 \usepackage{hyperref}
\usepackage{dsfont}
\usepackage{bm}
\usepackage{xcolor,soul,colortbl}
\usepackage{subcaption}
\usepackage{dashrule}

\newcommand{\mm}[1]{\ensuremath{\bm{#1}}} %Matrix (upper case)
\newcolumntype{C}[1]{>{\centering\arraybackslash}p{#1}}	
\newcolumntype{L}[1]{>{\raggedright\arraybackslash}p{#1}}
\newcolumntype{R}[1]{>{\raggedleft\arraybackslash}p{#1}}

\newcommand{\src}[1]{#1_s}
\newcommand{\trg}[1]{#1_t}
\newcommand{\cX}{\mathcal{X}}
\newcommand{\cY}{\mathcal{Y}}
\newcommand{\bx}{\mm x}
\newcommand{\by}{\mm y}
 
\newcolumntype{P}[1]{>{\centering\arraybackslash}p{#1}}

\newcommand{\thickhline}{%
    \noalign {\ifnum 0=`}\fi \hrule height 1pt
    \futurelet \reserved@a \@xhline
}

\begin{document}

\title{Domain Adaptation in LiDAR Semantic Segmentation \\by Aligning Class Distributions  }

\author{\authorname{I\~{n}igo~Alonso\sup{1},  Luis Riazuelo\sup{1}, Luis Montesano\sup{1,2} and Ana C. Murillo\sup{1}}
\affiliation{\sup{1}RoPeRt group, at DIIS - I3A, Universidad de Zaragoza, Spain}
\affiliation{\sup{2}Bitbrain, Zaragoza, Spain}
\email{\{inigo, riazuelo, montesano, acm\}@unizar.es}
}

\keywords{Robotics, Autonomous Systems, LiDAR, Deep Learning, Semantic Segmentation, Domain Adaptation}

\abstract{LiDAR semantic segmentation provides 3D semantic information about the environment, an essential cue for intelligent systems, such as autonomous vehicles, during their decision making processes. 
Unfortunately, the annotation process for this task is very expensive. To overcome this, it is key to find models that generalize well or adapt to additional domains where labeled data is limited. 
This work addresses the problem of unsupervised domain adaptation for LiDAR semantic segmentation models. 
We propose simple but effective strategies to reduce the domain shift by aligning the data distribution on the input space.  
Besides, we present a learning-based module to align the distribution of the semantic classes of the target domain to the source domain. 
Our approach achieves new state-of-the-art results on three different public datasets, which showcase adaptation to three different domains. 
}

\onecolumn \maketitle \normalsize \setcounter{footnote}{0} \vfill

\section{Introduction}

3D semantic segmentation is an important computer vision task that provides useful information to each registered 3D point of the surrounding environment. It has a wide range of applications and is particularly important in robotics, since most autonomous systems require an accurate and robust perception of their environment. LiDAR (Light Detection And Ranging) is a frequently used sensor for 3D perception in autonomous vehicles, such as cars or delivery robots, that provides accurate distance measurements of the surrounding 3D space. Despite LiDAR broad adoption, recognition systems such as semantic segmentation methods that generalize and perform well for different LiDAR sensors, vehicle set-ups or environments remains an unsolved challenging problem. Existing solutions typically need significant amounts of labeled data to adapt to different domains \cite{mei2019semantic}.

\begin{figure}[!tb]
\centering
\includegraphics[width=1\linewidth]{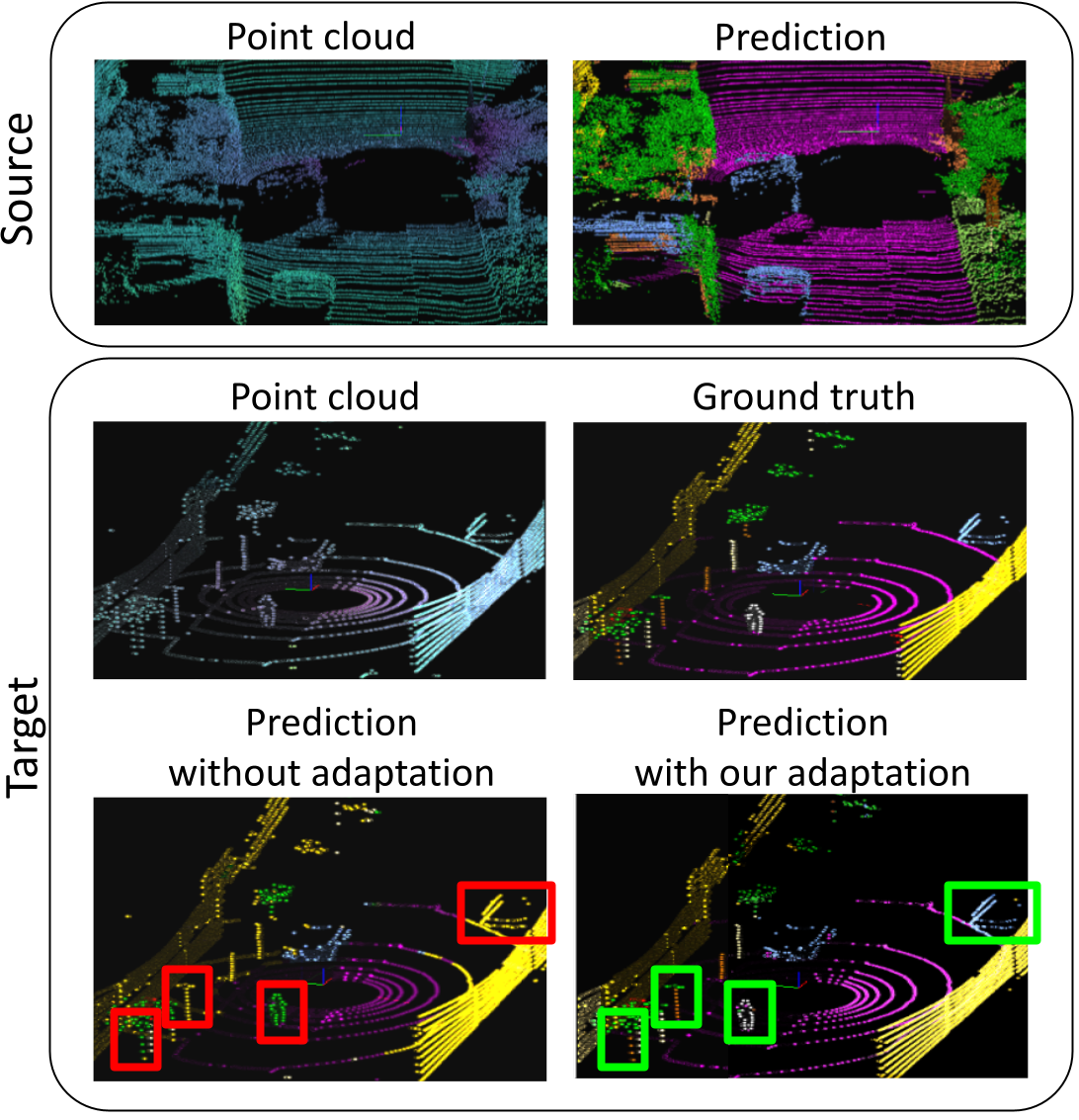}
\caption{Domain Adaptation in LiDAR Semantic Segmentation. Top row: segmentation result of a given model train on the same source domain \cite{behley2019semantickitti}. Bottom row: segmentation of the same model on a \textit{target domain} without adaptation and the improved result applying our proposed adaptation. Note relevant details missed if we do not adapt the model (marked with a red/green square).}  
\label{fig:introduction}
\end{figure} 

In recent years, deep learning methods are achieving state-of-the-art performance in 3D LiDAR semantic segmentation  \cite{milioto2019rangenet++, alonso20203d} when trained and evaluated in the same domain. Nevertheless, generalizing the learned knowledge to new domains or environments not seen during training is still a frequent open problem. When applying existing models on data with a different distribution than the training data, i.e., from a different domain, the performance is considerably degraded \cite{wang2018deep, tzeng2017adversarial, vu2019advent}. \\

Domain adaptation techniques aim to eliminate or reduce this drop. Existing works for domain adaptation in semantic segmentation focus on RGB data \cite{vu2019advent, zou2019confidence, chen2019domain, li2019bidirectional}. Most of them try to minimize the distribution shift between two different domains. Very few approaches have tackled this problem with LiDAR data \cite{wu2019squeezesegv2}, which equally suffers from the domain shift. 
RGB data commonly suffers from variations due to light and weather conditions, while the most common variations within 3D point clouds data come from sensor resolution (i.e., sensors with more laser sweeps generate denser point clouds) and from the sensor placement (because point clouds have relative coordinates with respect to the sensor). 
Both sensor resolution and placement issues are common examples that change the data distribution of the captured 3D point clouds. Coping with these issues would enable the use of large existing labeled LiDAR datasets for more realistic use-cases in robotic applications, reducing the need for data labeling.

This work proposes two strategies to improve unsupervised domain adaptation (UDA) in LiDAR semantic segmentation, see a sample result on Fig. \ref{fig:introduction}.  The first strategy addresses this problem by applying a set of simple data processing  steps to align the data distribution reducing the domain gap on the input space.
The second strategy proposes how to align the distribution on the output space by aligning the class distribution. 
These two proposed strategies can be applied in conjunction with current state-of-the-art approaches boosting their performance.

We validate our approach on  different scenarios getting state-of-the-art results. We use the SemanticKitti \cite{behley2019semantickitti} as the source domain and we adapt it to SemanticPoss \cite{pan2020semanticposs}, to Paris-Lille-3D \cite{roynard2018paris} and to a new collected and released dataset.

\section{Related Work}
This section describes the related work for the most relevant related topics for this work: 3D LiDAR segmentation and,  unsupervised domain adaptation for semantic segmentation.

\subsection{3D LiDAR Point Cloud Segmentation}
3D LiDAR Semantic segmentation  aims to assign a semantic label to every point scanned by the LiDAR sensor.
Before the current trend and wide adoption of deep learning approaches, earlier methods relied on exploiting prior knowledge and geometric constraints \cite{xie2019review}.
As far as deep learning methods are concerned, there are two main types of approaches to tackle the 3D LiDAR semantic segmentation problem. On one hand, there are approaches that work directly on the 3D points, i.e., the raw point cloud is taken as the input \cite{qi2017pointnet, qi2017pointnet++}. On the other hand, other approaches convert this 3D point cloud into another representation (images\cite{alonso20203d}, voxels\cite{zhou2018voxelnet}, lattices\cite{rosu2019latticenet})  in order to have a structured input. For LiDAR semantic segmentation, the most commonly used representation is the spherical projection \cite{alonso20203d, milioto2019rangenet++, wu2018squeezeseg}.
Several recent works \cite{milioto2019rangenet++, wu2018squeezeseg, alonso20203d} show that point-based methods, i.e., approaches that work directly on the 3D points, are slower and tend to be less accurate than methods which project the 3D point cloud into a 2D representation and make use of convolutional layers.

\subsection{Unsupervised Domain Adaptation for Semantic Segmentation}

Unsupervised Domain Adaptation (UDA) aims to adapt models that have been trained on one specific domain (source domain) to be able to work on a different domain  (target domain) where there is a certain lack of labeled training data. 
Most works follow similar ideas: the input data or features from a source-domain sample and a target-domain sample should be indistinguishable. 
Several works follow an adversarial training scheme to minimize the distribution shift between the target and source domains data. This approach has been shown to work properly at pixel space \cite{yang2020fda}, at feature space \cite{hoffman2018cycada} and at output space \cite{vu2019advent}.
However, adversarial training schemes, such as DANN \cite{ganin2016dann}, tend to present convergence problems. Alternatively, other works follow different schemes like CORL \cite{sun2017correlation} or MinEnt \cite{vu2019advent}. Entropy minimization methods \cite{vu2019advent, chen2019domain} do not require complex training schemes. They rely on a loss function that minimizes the entropy of the unlabeled target domain output probabilities.
 
% NOW 3D/LiDAR DOMIAN ADAPTAITON
Regarding segmentation on LiDAR data, very few works have studied the problem of domain adaptation.  
 SqueezeSegV2 \cite{wu2019squeezesegv2} proposes how to adapt synthetic data where only coordinates are available without reflectance to real data. For this, they propose a learned intensity rendering to create the intensity for the synthetic data. 
A very recent work, Xmuda \cite{jaritz2020xmuda} focuses on combining different modalities: LiDAR and RGB for multi-modal domain adaptation. They propose to apply the KL divergence between the output probabilities of both modalities as the main loss function. Besides, they also apply previously proposed methods like entropy minimization \cite{vu2019advent}.
Differently, this work investigates different UDA strategies (both existing and novel) to improve UDA for the particular case of urban LiDAR semantic segmentation. The presented results show their effectiveness in reducing the domain gap.

\section{Unsupervised Domain Adaptation For LiDAR Semantic Segmentation}
\label{sec:method}

\begin{figure*}[!tb]
\centering
\includegraphics[width=0.95\linewidth]{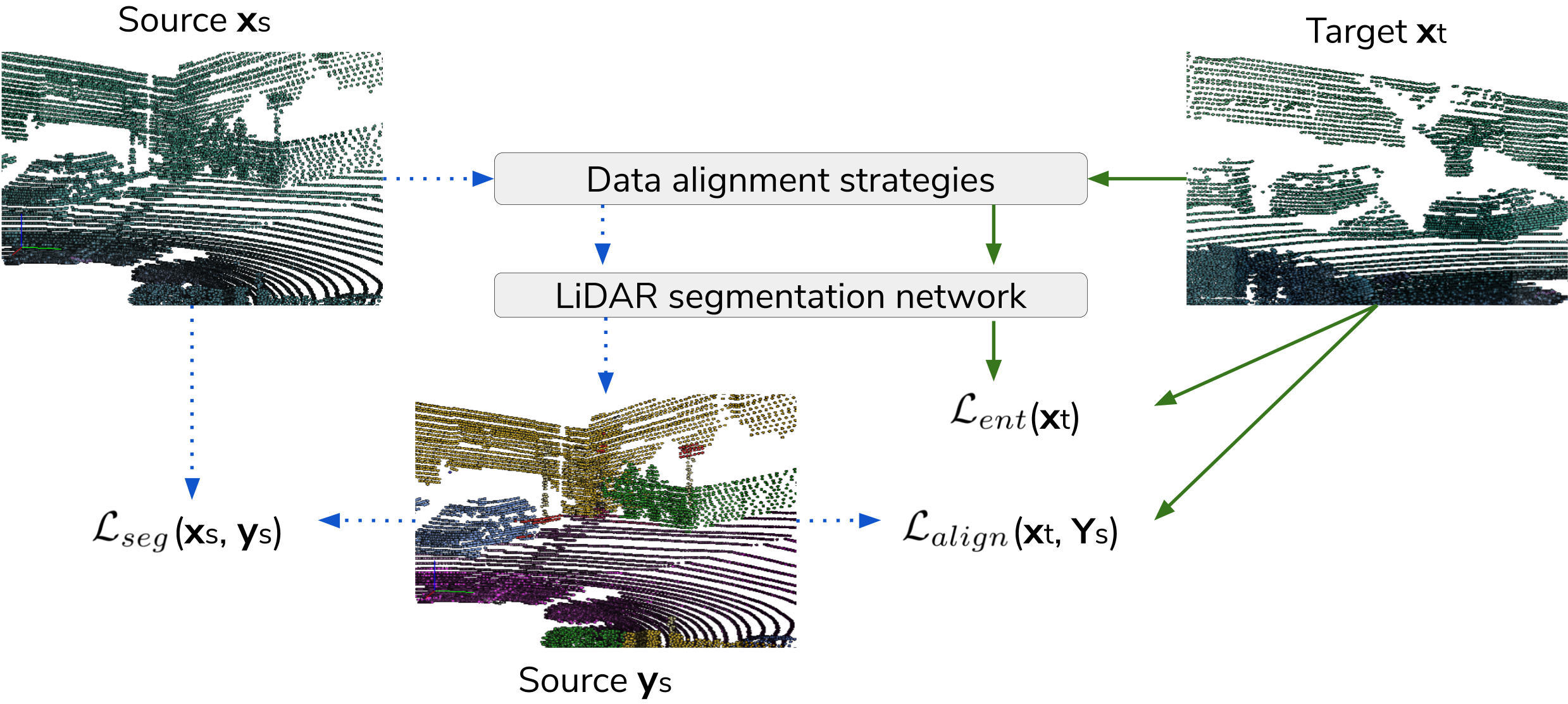}
\caption{\textbf{Approach overview.} The figure shows our pipeline steps and optimization losses. First, we perform distribution alignment on the input space, i.e., data alignment strategies. Then, we optimize the segmentation loss for source samples where the labels are known and the class alignment and entropy losses for target data where no labels are available (See Sect. \ref{sec:method} for details).
 \textcolor{Green}{Green continuous arrows} are used for target data and \textcolor{Blue}{blue dotted arrows}  for source data.}  
\label{fig:pipeline}
\end{figure*}

This section describes the proposed domain adaptation approach, including the LiDAR semantic segmentation method used, the strategies proposed to reduce the domain gap (data alignment and class distribution alignment), and the formulation of the proposed learning task.   
Figure \ref{fig:pipeline} presents an overview of our proposed approach which is further explained in the following subsections.

\subsection{LiDAR Semantic Segmentation Model}
\label{sec:LiDAR-segmentation} 
We use a recent method for LiDAR semantic segmentation which achieves state-of-the-art performance on several LiDAR segmentation datasets, 3D-MiniNet \cite{alonso20203d}. 
This method consists of three main steps. First, it learns a 2D representation from the 3D points. Then, this representation is fed to a 2D Fully Convolutional Neural Network that produces a 2D semantic segmentation. These 2D semantic labels are re-projected back to the 3D space and enhanced through a post-processing module.

Let $\src{\cX}\subset \mathbb{R}^{N\times 3}$ be a set of source domain LiDAR point clouds along with associated semantic labels, $\src{\cY}\subset (1,C) ^{N}$.
Sample $\bx_s$ is a point cloud of size $N$ and $ \by_s^{(n,c)} $ provides the label of point $(n)$  as one-hot encoded  vector. 
Let $F$ be our LiDAR segmentation network which takes a point cloud $\mm x$ and predicts a  probability distribution (size $C$ classes) for each point of the point-cloud $F(\bx) =  \mm P_{\mm x}^{(n,c)}$.

The parameters $\theta_F$ of $F$ are optimized  to minimize cross-entropy loss:
\begin{equation}
\label{eq:onlysource}
\mathcal{L}_{seg}(\src{\mm x}, \src{\mm y}) = -\sum_{n=1}^N\sum_{c=1}^{C} \src{\mm y}^{(n,c)} \log \mm P_{\src{\mm x}}^{(n,c)},
\end{equation}
on source domain samples. Therefore, as the supervised semantic segmentation is concerned, the optimization problem simply reads:
\begin{equation}
\label{eq:onlysource}
\min_{\theta_F} \frac{1}{|\cX_s|}\sum_{\mm x_s\in\cX_s} \mathcal{L}_{seg}(\src{\mm x}, \src{\mm y}).
\end{equation}

\subsection{Data alignment strategies for LiDAR} 
\label{sec:strategies}

The problem of domain adaptation, i.e., data distribution misalignment, between $\src{\cX}$ and $\trg{\cX}$ (a set of target domain LiDAR point clouds), can be handled on the network weights  ${\theta_F}$ but also modifying $\src{\cX}$ and $\trg{\cX}$ in order to align the distributions at the input space.

Next, we describe the different strategies for better data alignment that we propose to improve LiDAR domain adaptation.

\paragraph*{\textbf{XYZ-shift augmentation}} One of the main causes of misalignment for LiDAR point clouds is the location of the sensor.
Since the point cloud values are relative to the sensor origin, these changes cause variations affecting the whole point cloud. We perform shifts up to $\pm$2 meters on the Z-axis (height) and up to $\pm$5 meters on the Y-axis and X-axis.
    
\paragraph*{\textbf{Per-class augmentation}} We also propose to perform the augmentation independently per class, i.e., applying different data augmentation parameters for every class. In particular, in this work, we perform shifts up to $\pm$1 meters on the Z-axis (height) and up to $\pm$3 meters on the Y-axis and X-axis.

\paragraph*{\textbf{Same number of beams}} Besides the sensor placement and orientation, a significant difference between sensors is the number of captured beams, which results in a more sparse or dense point cloud. We propose to match the data beams between the two domains by reducing the data from the sensor with a higher number of beams ending up with more homogeneous data within  $\src{\cX}$ and  $\trg{\cX}$. 

\paragraph*{\textbf{Only relative features}} Point-cloud segmentation methods commonly use both absolute and relative (between points) values of the input data. In order to be independent of absolute coordinates that are less robust compared to relative coordinates, we propose to use only relative features of the data. Therefore, we propose to use only relative distances of every point with respect to their neighbors.

\subsection{Class distribution alignment}
The domain shift appears due to many different factors. For example, different environments can present quite different appearances, the spatial distribution of objects may vary, the capturing set-up for different scenarios can be totally different, etc. 
Depending on the problem tackled and prior knowledge, we can hypothesize which of these differences can be neglected and assumed not to affect to the models we are learning.
In this work, we tackle the domain adaptation problem between different urban environments, since all the datasets used are from urban scenarios.
Taking this into account, although the data distribution changes between the datasets, we can assume that the class distribution is going to be very similar across these scenarios since the distribution shift in urban environments is mostly due to appearance and  spatial distribution changes. For example, we can assume that if $\src{\mm{y}}$ has a distribution of 90\% road pixels and 10\% car pixels, then $\trg{\mm{y}}$ will likely present a similar distribution.  

Our approach learns parameters $\theta_F$ of $F$ in such a way that the predicted class distribution  $F(\trg{\cX})$ matches the real class distribution of $\src{\cY}$, i.e., the histogram representing the frequency of occurrence of each class, previously computed in an offline fashion.
To do so, We propose to compute the KL-divergence between these two class distribution as the class alignment loss 
\begin{equation}
\label{eq:onlysource}
\mathcal{L}_{align}(\trg{\mm x}, \src{\cY}) = \sum_{n=1}^N  \mm hist(\src{\cY}) \log\frac{ \mm  hist(\src{\cY})}{ \mm P_{\trg{\mm x}}^{(n)}}.
\end{equation}
Therefore, the optimization problem reads as:
\begin{equation}
\label{eq:clasdistribution}
\min_{\theta_F} \frac{1}{|\cX_t|}\sum_{\mm x_t\in\cX_t} \mathcal{L}_{align}(\trg{\mm x}, \src{\cY}). 
\end{equation} 

Equation 2 requires to compute  the class distribution $\mm P_{\mm x_t}$ over the whole dataset. As this is computationally unfeasible, we compute it over the batch as an approximation.

\subsection{Optimization Formulation}

The entropy loss is computed as in MinEnt \cite{vu2019advent}:
\begin{equation}\label{eq:minentloss}
\mathcal{L}_{ent}(\trg{\mm x}) =\frac{-1}{\log(C)}\sum_{n=1}^N\sum_{c=1}^{C} \mm P_{\trg{\mm x}}^{(n,c)} \log \mm P_{\trg{\mm x}}^{(n,c)},
\end{equation}
while the segmentation $\mathcal{L}_{seg}$ and class alignment   $\mathcal{L}_{align}$ losse are computed as detailed in previous subsections.

During training, we jointly optimize the supervised segmentation loss $\mathcal{L}_{seg}$ on source samples and the class alignment loss  $\mathcal{L}_{align}$ and entropy loss $\mathcal{L}_{ent}$ on target samples. 
The final optimization problem is formulated as follows:

\begin{equation}\label{eq:losses}
\begin{split}
&\min_{\theta_F} \frac{1}{|\cX_s|}\sum_{\bx_s} \mathcal{L}_{seg}(\src{\mm x}, \src{\mm y})  +  \frac{\lambda_{align}}{|\cX_t|} \sum_{\bx_t}\mathcal{L}_{align}(\trg{\mm x}, \src{\cY}) \\   
&+  \frac{\lambda_{ent}}{|\cX_t|} \sum_{\bx_t}\mathcal{L}_{ent}(\trg{\mm x}),
\end{split}
\end{equation}
with~$\lambda_{ent}$ and $\lambda_{align}$ as the weighting factors of the alignment and entropy terms.

\section{Experimental Setup}
This section details the setup used in our evaluation. This includes the datasets used in the evaluation and the training protocol we followed.

\subsection{Datasets}
\label{sec:datasets}
We use four different datasets for the evaluation. 
They were collected in four different geographical areas, with four different LiDAR sensors, and with four different set-ups. 
We take the well known SemanticKITTI dataset \cite{behley2019semantickitti} as the source domain dataset and the other three datasets as target domain data.

\paragraph*{SemanticKITTI}
The SemanticKITTI dataset \cite{behley2019semantickitti} is a recent large-scale dataset that provides dense point-wise annotations.
The dataset consists of over 43000 LiDAR scans from which over 21000 are available for training. The dataset distinguishes 22 different semantic classes. The capturing sensor is a Velodyne HDL-64E mounted on a car.

\paragraph*{Paris-Lille-3D}
Paris-Lille-3D \cite{roynard2018paris} is a medium-size dataset that provides three aggregated point clouds. It is collected with a tilted rear-mounted Velodyne HDL-32E placed on a vehicle. Following PolarNet work \cite{zhang2020polarnet}, we extract individual scans from the registered point clouds thanks to the scanner trajectory and points’ timestamps. Each scan is made of points within +/- 100m. We take the Lille-1 point cloud for the domain adaptation and Lille-2 for validation. 
We use the following intersecting semantic classes with the SemanticKitti: car, person, road, sidewalk, building, vegetation, pole, and traffic light.

\paragraph*{SemanticPoss}
The SemanticPoss  \cite{pan2020semanticposs} is a medium-size dataset which contains 5 different sequences from urban scenarios providing 3000 LiDAR scans. The sensor used is a 40-line Pandora mounted on a vehicle. 
We take the three first sequences for applying the adaptation methods and the last two sequences for validation.
We use the following intersecting classes with the SemanticKitti: car, person, trunk, vegetation, traffic sign, pole, fence, building, rider, bike, and ground (combines road and sidewalk).

\paragraph*{I3A} 
We have captured a small dataset to test our approach in a different scenario. In contrast to the three previous datasets, this dataset is not captured from a vehicle but from a TurtleBot, having the sensor at a lower height than in the other set-ups.
The capturing sensor is the Velodyne VLP-16. We capture an urban environment similar to previous datasets.   
The dataset contains two sequences, one for training and another for validation.  
We use the intersecting semantic classes with the SemanticKitti: car, person, road, sidewalk, building, vegetation, trunk, pole, and traffic light.

\subsection{Training Protocol}
As we mentioned in Sec. \ref{sec:LiDAR-segmentation}, we use 3D-MiniNet \cite{alonso20203d} as the base LiDAR semantic segmentation method. In particular, we use the available 3D-MiniNet-small version because of memory issues. 
For computing the relative coordinates and features, we follow 3D-MiniNet approach extracting them from the N neighbors of each 3D point where N is set to 16. 

For all the experiments we train this architecture for 700K iterations with a batch size of 8.   
We use Stochastic Gradient Descent (SGD) optimizer with an initial learning rate of $0.005$ and a polynomial learning rate decay schedule with a power set to 0.9. 

We set $\lambda_{ent}$ to 0.001 as suggested in MinEnt \cite{vu2019advent} and $\lambda_{align}$ to 0.001. We empirically noticed that the performance is very similar when these two hyper-parameters are set between $10^{-5}$ and  $10^{-2}$. The two main conditions for them to work properly are: (1) be greater than 0 and, (2) do not be higher than the supervised loss.

One thing to take into account is that, as explained in \ref{sec:datasets}, the Paris-Lille-3D has a very limited field of view. Therefore in order to make MinEnt \cite{vu2019advent} work in this dataset, we had to simulate the same field of view on the source dataset. 

\section{Results} 

This section presents the experimental validation of our approach compared to different baselines. The proposed approach achieves better results than the other baselines in the three different scenarios for unsupervised domain adaptation in LiDAR Semantic Segmentation. In all the experiments we use the SemanticKITTI dataset \cite{behley2019semantickitti} as the source data distribution and perform the adaptation on the other three datasets.

\begin{table}[!tb]
\centering
\caption{
Ablation study of our domain adaption pipeline for semantic segmentation. Source dataset: SemanticKitti \cite{behley2019semantickitti}.}
\label{tab:ablation}
\resizebox{\columnwidth}{!}{
\begin{tabular}{lccc}
\toprule[1.0pt]
 & \textbf{mIoU on} & \textbf{mIoU on} & \textbf{mIoU on}  \\
Target dataset &  \textbf{I3A} &\textbf{ParisLille} & \textbf{SemanticPoss}  \\%  \rotatebox{90}{\textbf{mIoU}}
\toprule[1.0pt]
Base model & 15.9  & 19.2 & 14.0\\
+ XYZ-shift augmentation & 25.1 & 28.9 &16.8 \\
+ Per-class augmentation &  27.0  &30.1 & 18.2  \\
+ Same number of beams &  42.0  &35.4 & 19.7  \\
+ Only relative features &  47.1  & --- & 21.8  \\
+ MinEnt \cite{vu2019advent}  &  50.3  &41.5 & 26.2  \\
+ Class distrib. alignment  &  52.5  &42.7 & 27.0  \\
\toprule[1.0pt]
   \multicolumn{4}{p{8cm}}{\footnotesize {--- Not used because there was no performance gain.}}
\end{tabular} } 
\end{table}

\begin{figure*}
\centering
\begin{tabular}{c}
     \includegraphics[width=0.86\linewidth]{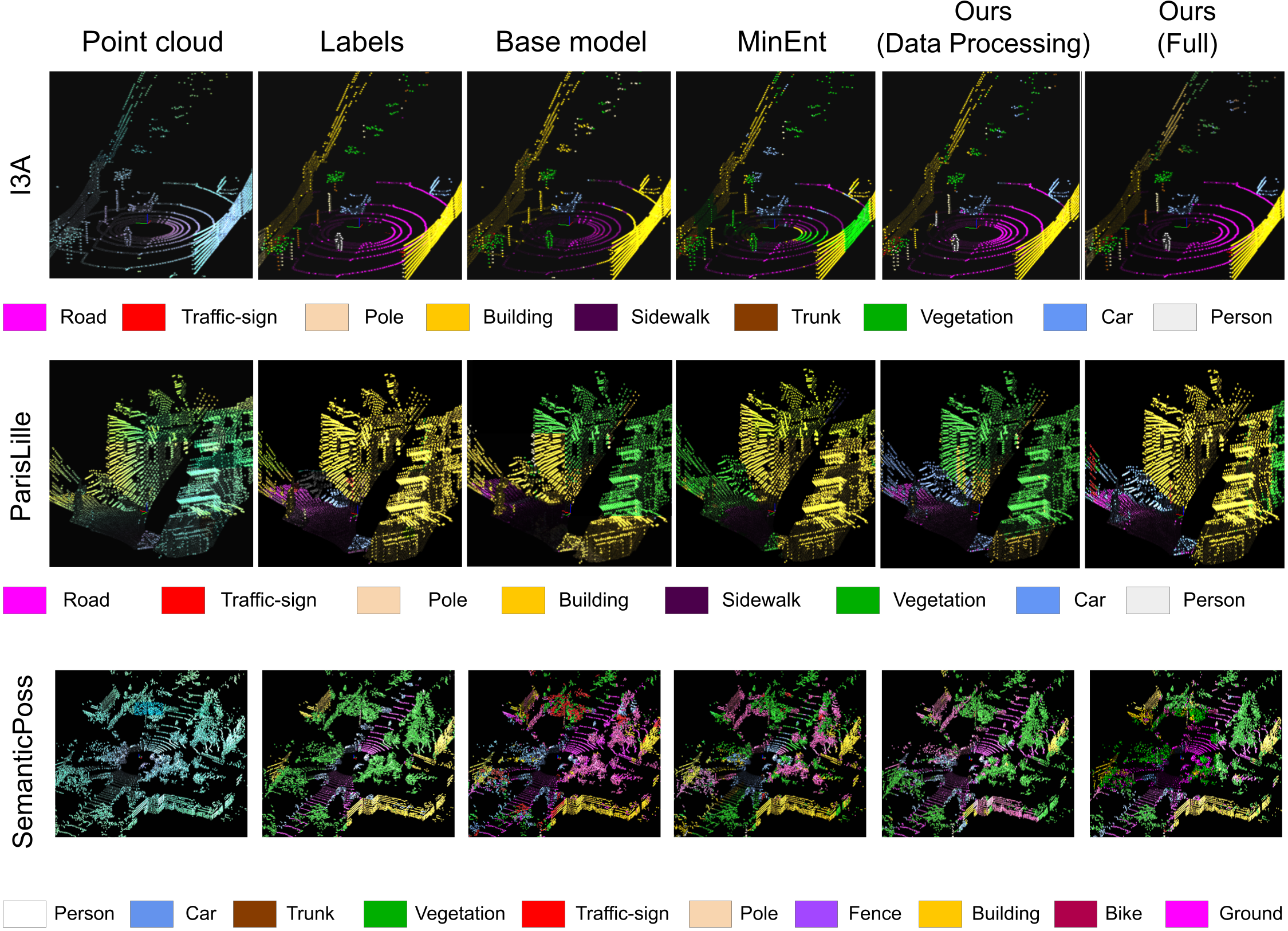}
\end{tabular}
\caption{LiDAR domain adaptation results for different methods and datasets: I3A dataset  first row, ParisLille dataset second row, and SemanticPoss last row. From left to right: Input point cloud, ground truth labels, baseline with no adaptation (trained on SemanticKitti), MinEnt \cite{vu2019advent}, our adaptation only with data processing strategies, our full adaptation pipeline. Best viewed in color. For better and more detail (video result) go to \url{https://youtu.be/EpkJ_UH1F-o}.} \vspace{-3mm}
\label{fig:visual-results}
\end{figure*} 

\subsection{Ablation Study}

The experiments in this subsection show how the different data alignment steps and the proposed learning losses  affect the final performance of our approach. Table \ref{tab:ablation} summarizes the  ablation study performed on three different scenarios. The results show how all the steps proposed contribute towards the final performance. The main insights observed in the ablation study are discussed next.

Performing strong \textit{XYZ-shifts} results in a boost on the performance, meaning that the domain gap is considerably reduced. The distribution gap reduced by this step is the one caused by the fact of using different LiDAR sensor set-ups (such as different acquisition sensor height). Besides, in these autonomous driving set-ups, the distance between the car and the objects depends on how wide are the streets or on which lane is the data capturing source. Therefore, this is an essential and really easy data transformation to perform which gives an average of 7.2\% MIoU gain.

The \textit{per-class data augmentation} also boosts the performance. This data augmentation method tries to reduce the domain gap by adding different relative distances between different classes gaining an average of 1.3\% MIoU gain.

Another interesting and straightforward technique to perform is to \textit{match the number of LiDAR beams} of the source and target data, i.e., match LiDAR point-cloud resolution. This helps the data alignment especially for having the same point density on the 3D point-cloud and similar relative distances between the points. We show that this method gives an improvement similar to the XYZ-shift data augmentation, hugely reducing the domain gap. The higher the initial difference in the number of beams, the more improvement we can get: the i3A LiDAR has 16 beams, the ParisLille 32, and the SemanticPoss 40, compared to the 64 of the source data (SemanticKitti).

The use of \textit{relative features only} does not always help to reduce the domain gap, it was only beneficial on the i3A and SemanticPoss datasets.  Removing the absolute features and only learning from relative features helps especially when the relative distances between the 3D points have less domain shift than the absolute coordinates. This will depend on the dataset, but the stronger the differences between capturing sensors, the more likely that the use of relative features will help.

Besides the data alignment steps, our approach includes two \textit{learning losses} to the pipeline to help to reduce the domain gap. The first one is the entropy minimization loss proposed in MinEnt \cite{vu2019advent}. The second one is our proposed class distribution alignment loss introduced in this work. We show that these two losses can be combined for the domain adaptation problem and that, although less significantly with respect to previously discussed steps, they also improve on the three different set-ups, contributing to achieving state-of-the-art performance.

\subsection{Comparison with other baselines}

\begin{table}[!bh]
\centering
\caption{
Results on the three different LiDAR semantic segmentation datasets using different domain adaptation methods. The source dataset is the SemanticKitti dataset \cite{behley2019semantickitti} }
\label{tab:adaptation}

\resizebox{\columnwidth}{!}{
\begin{tabular}{rccc}
\toprule[1.0pt]
 & \textbf{mIoU on} & \textbf{mIoU on} & \textbf{mIoU on}  \\
 &  \textbf{I3A} &\textbf{ParisLille} & \textbf{SemanticPoss}  \\
\toprule[1.0pt]
Baseline & 15.9  & 19.2 & 13.4\\
MinEnt \cite{vu2019advent} & 28.4  & 23.2 & 19.6 \\
AdvEnt \cite{vu2019advent} & 21.0 & 20.7 & 19.5\\
MaxSquare \cite{chen2019domain} & 28.4 & 22.8 &19.3\\
\toprule[1.0pt]
Data alignment (ours)* & 47.1 & 36.2 &19.0 \\
Full approach (ours) &  52.5  &42.7 & 27.0  \\
\toprule[1.0pt]
   \multicolumn{4}{p{8cm}}{\footnotesize {* Only data alignment strategies from Sect. \ref{sec:strategies} }}

\end{tabular} }

\end{table}

Table \ref{tab:adaptation} shows the comparison of our pipeline (composed of all the steps discussed in the ablation study) with other existing methods for domain adaptation. 
We select MinEnt, Advent, and MaxSquare as the baselines because they are leading the state-of-the-art for unsupervised domain adaptation. We use the available authors' code for replication.

We apply the different domain adaptation methods of the three different set-ups without our data alignment steps. This comparison shows that good pre-processing of the data can obtain better results than just applying out-of-the-box methods for domain adaptation. 
It also shows that our complete pipeline outperforms these previous methods on LiDAR domain adaptation.
Our results demonstrate that combining proper data processing with learning methods for domain adaptation gives an average of more than $\times2$ boost on the performance.

Figure \ref{fig:visual-results} includes a few examples of the segmentation obtained with a baseline with no domain adaptation, using the MinEnt \cite{vu2019advent} approach only, with our approach using only the data pre-processing steps, and with our approach including all steps proposed. We can appreciate in figure \ref{fig:visual-results} how data processing helps on certain semantic classes, such as road, person, car, or vegetation, while MinEnt usually improves at different ones like building. This suggests the good complementary of both strategies, and indeed combining them provides the best results. 
More detail in additional video results can be seen at \url{https://youtu.be/EpkJ_UH1F-o}.

\section{Conclusions}
In this work, we introduce a novel pipeline that addresses the task of unsupervised domain adaptation for LiDAR semantic segmentation. Our pipeline consists of aligning data distributions on the data space with different simple strategies combined with learning losses on the semantic segmentation process that also force the data distribution alignment.
Our results show that a proper data alignment on the input space can produce better domain adaptation results that just using out-of-the-box state-of-the-art learning methods. Besides, we show that combining these data alignment methods with learning methods, like the one proposed in this work to align the class distributions of the data, can reduce even more the domain gap getting new state-of-the-art results. Our approach is validated on three different scenarios, from different datasets, as the target domain, where we show that our full pipeline improves previous methods on all three scenarios.

\section{Acknowledgments}
This project was partially funded by projects FEDER/ Ministerio de Ciencia, Innovaci{\'o}n y Universidades/ Agencia Estatal de Investigaci{\'o}n/RTC-2017-6421-7, PGC2018-098817-A-I00 and PID2019-105390RB-I00, Arag{\'o}n regional government (DGA T45 17R/FSE) and the Office of Naval Research Global project ONRG-NICOP-N62909-19-1-2027.

{
\bibliographystyle{apalike}
\bibliography{biblio}
}

\end{document}